\documentclass[11pt]{article}
\usepackage{syntaxfest2021}
\usepackage{times}
\usepackage{url}
\usepackage{latexsym}
\usepackage{multirow}
\usepackage{enumitem}
\usepackage{hyperref}

\usepackage{graphicx}
\usepackage{siunitx}
\usepackage{float}

\usepackage[title]{appendix}

\usepackage{pgffor}

\syntaxfestfinalcopy % Uncomment this line for the final submission

\newcommand*{\FirstIndent}{\hspace*{0.5cm}}%

\usepackage{fancyhdr}
\title{Minor changes make a difference: a case study on the consistency of UD-based dependency parsers}

\author{Dmytro Kalpakchi \\
  Division of Speech, Music and Hearing \\
  KTH Royal Institute of Technology \\
  Stockholm, Sweden \\
  \texttt{dmytroka@kth.se} \\\And
  Johan Boye \\
  Division of Speech, Music and Hearing \\
  KTH Royal Institute of Technology \\
  Stockholm, Sweden \\
  \texttt{jboye@kth.se} \\}

\date{}

\begin{document}
\maketitle
\thispagestyle{fancy}
\begin{abstract}
Many downstream applications are using dependency trees, and are thus relying on dependency parsers producing correct, or at least consistent, output. However, dependency parsers are trained using machine learning, and are therefore susceptible to unwanted inconsistencies due to biases in the training data. This paper explores the effects of such biases in four languages -- English, Swedish, Russian, and Ukrainian -- though an experiment where we study the effect of replacing numerals in sentences. We show that such seemingly insignificant changes in the input can cause large differences in the output, and suggest that data augmentation can remedy the problems.

\end{abstract}

%
% The following footnote without marker is needed for the camera-ready
% version of the paper.
% Comment out the instructions (first text) and uncomment the 8 lines
% under "final paper" for your variant of English.
% 
\begin{NoHyper}
\blfootnote{
    %
    % for review submission
    %
    %\hspace{-0.65cm}  % space normally used by the marker
    %Place licence statement here for the camera-ready version.

    % final paper: en-uk version 
    %
    \hspace{-0.65cm}  % space normally used by the marker
    This work is licensed under a Creative Commons 
    Attribution 4.0 International Licence.
    Licence details:
    \url{http://creativecommons.org/licenses/by/4.0/}.
    % 
    % % final paper: en-us version 
    %
    % \hspace{-0.65cm}  % space normally used by the marker
    % This work is licensed under a Creative Commons 
    % Attribution 4.0 International License.
    % License details:
    % \url{http://creativecommons.org/licenses/by/4.0/}.
}
\end{NoHyper}

\section{Introduction}
\label{intro}
The Universal Dependencies (UD) resources have steadily grown over the years, and now treebanks for over 100 languages are available. The UD community has made a tremendous effort in providing a rich toolset for utilizing the treebanks for downstream applications, including pre-trained models for dependency parsing \cite{straka-etal-2016-udpipe,qi-etal-2020-stanza} and tools for manipulating UD trees \cite{popel-etal-2017-udapi,peng-zeldes-2018-roads,kalpakchi-boye-2020-udon2}.

Such an extensive infrastructure makes it more appealing to develop multilingual downstream applications based on UD, as a deterministic and more explainable competitor to the currently dominant neural methods. It is also compelling to use UD-based metrics for evaluation in multilingual settings. In fact, researchers have already started exploring such possibilities on both mentioned tracks. Kalpakchi and Boye \shortcite{kalpakchi2021quinductor} proposed a UD-based multilingual method for generating reading comprehension questions. Chaudhary et al. \shortcite{chaudhary-etal-2020-automatic} designed a UD-based method for automatically extracting rules governing morphological agreement. Pratapa et al. \shortcite{pratapa2021evaluating} proposed a UD-based metric to evaluate the morphosyntactic well-formedness of generated texts. 

The authors of the latter two articles trained their own more robust versions of the dependency parsers, suitable for their needs. The authors of the first article relied on the off-the-shelf model, making the robustness of pre-trained dependency parsers crucial for the success of the downstream applications. For instance, sentence simplification rules based on dependency trees might simply not fire due to a mistakenly identified head or dependency relation. In fact, state-of-the-art dependency parsers are somewhat error-prone and not perfect, and assuming otherwise might potentially harm the performance of downstream applications. A more relaxed (and realistic) assumption is that the errors made by the parser are at least {\em consistent\/}, so that potentially useful patterns for the task at hand can still be inferred from data. These patterns might not always be linguistically motivated, but if the dependency parser makes consistent errors, they can still be useful for the task at hand.

% Useful for negative sampling when training ML methods
In this article, we perform a case study operating under this relaxed assumption and investigate the consistency of errors while parsing sentences containing numerals. This step is useful, for instance, in question generation (especially for reading comprehension in the history domain) or numerical entity identification (e.g., distinguishing years from weights or distances).

\section{Background: Convolution partial tree kernels}
%Convolution partial tree kernels (CPTKs) were originally proposed by Moschitti \shortcite{moschitti2006efficient} for dependency trees for calculating the number of common substructures between two given trees. The basic idea is to represent trees as vectors in a common vector space and name a dot product between these vectors to be CPTK (as illustrated in Figure \ref{fig:cptk_example}). However, the vector space is induced only implicitly, whereas CPTK itself is calculated using a dynamic programming algorithm (for more details we refer to the original article).

In order to measure parser accuracy, metrics like Unlabelled or Labelled Attachment Score (UAS and LAS, respectively) are often used. However, these metrics they do not fully reflect the usefulness of the parsers in downstream applications. A minor error in attaching one dependency arc will result in a minor decrease in UAS and LAS. In fact, the very same minor error might lead to a completely unusable tree for the task at hand, depending on how close the error is to the root. Therefore, we need a metric that penalizes errors more the closer the errors are to the root. 

%In fact, even a minor error in attaching one dependency arc might lead to a completely wrong tree for the task at hand, depending on how close the error is to the root. Therefore, we need a metric that penalizes errors more the closer the errors are to the root. 

\begin{figure}
    \centering
    \begin{minipage}{.68\textwidth}
        \centering
        \includegraphics[width=\textwidth]{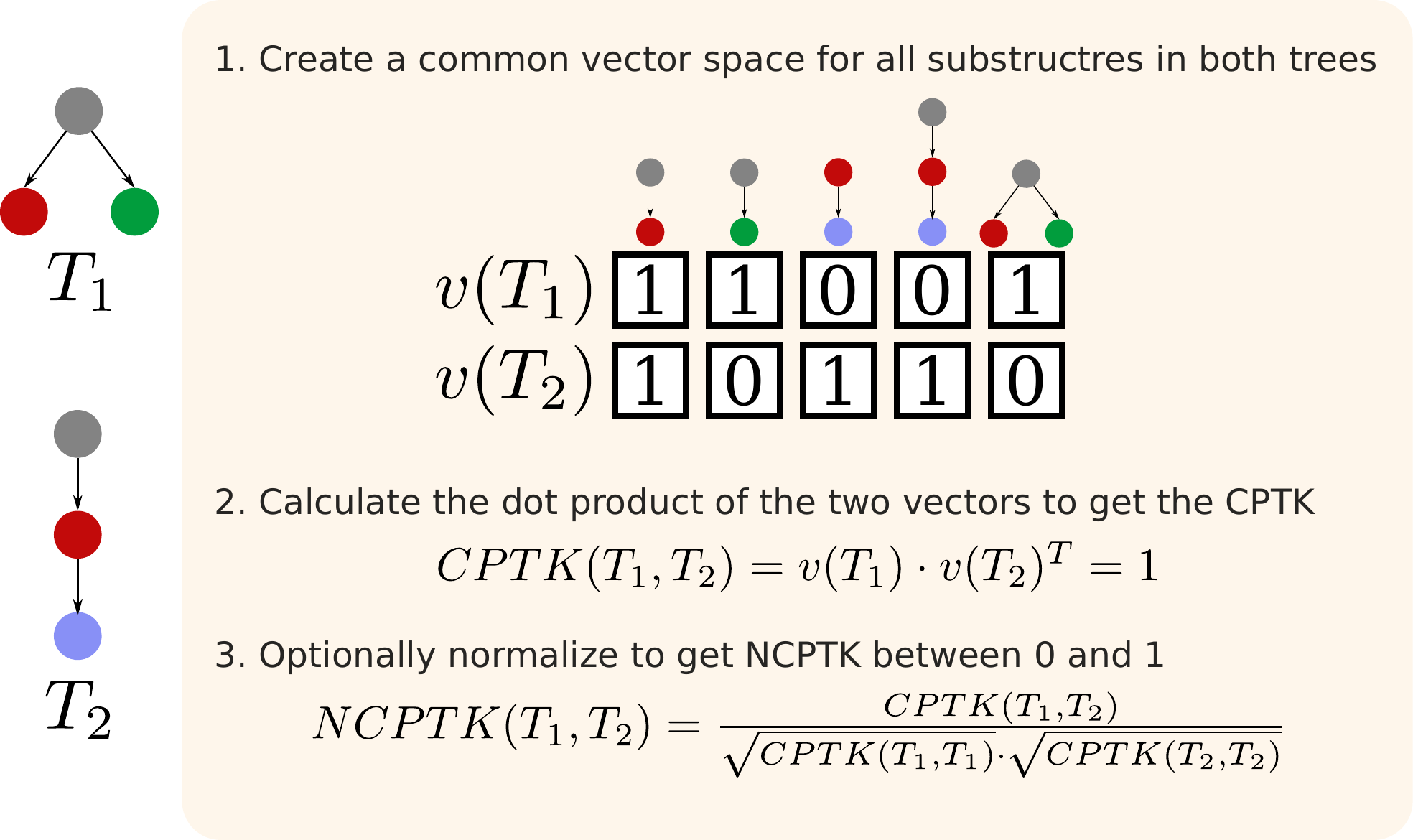}
    	\caption{A simple example illustrating \emph{the concept} behind convolution partial tree kernels (in practice the vector space is induced only implicitly and CPTK is calculated using dynamic programming)}
    	\label{fig:cptk_example}
    \end{minipage}%
    \hspace{1.3em}
    \begin{minipage}{0.28\textwidth}
        \centering
        \includegraphics[width=\textwidth]{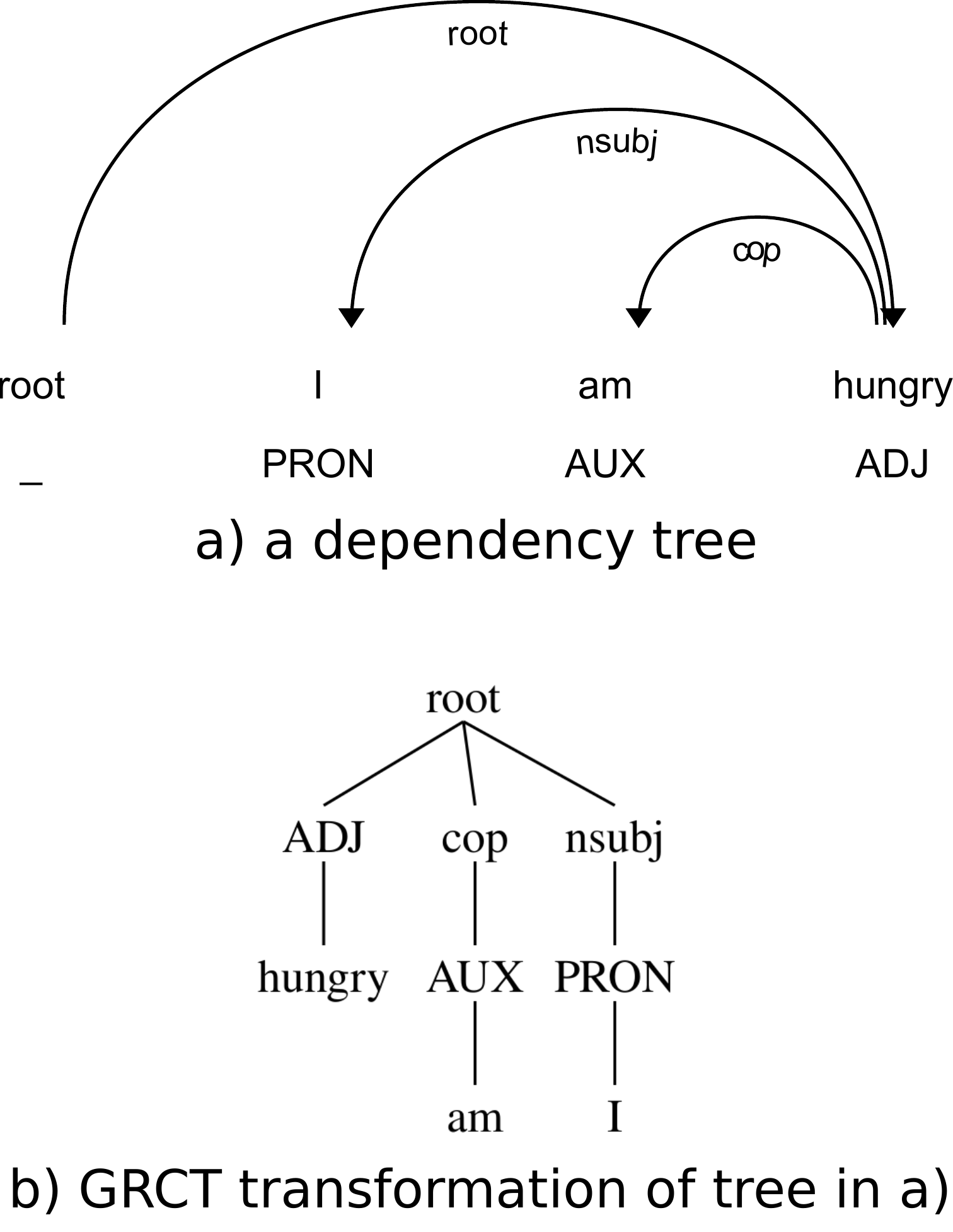}
    	\caption{A simple example of a GRCT transformation}
    	\label{fig:grct_example}
    \end{minipage}
\end{figure}

One metric possessing this desirable property is the convolution partial tree kernel (CPTK), originally proposed by Moschitti \shortcite{moschitti2006efficient} as a similarity measure for dependency trees. The basic idea is to represent trees as vectors in a common vector space, in such a way that the more common substructures two given trees have, the higher the dot product is between the corresponding two vectors (as illustrated in Figure~\ref{fig:cptk_example}). However, the vector space is induced only implicitly, whereas the dot product (the CPTK) itself is calculated using a dynamic programming algorithm (for more details we refer to the original article). CPTK values increase with the size of the trees, and thus can take any non-negative values, making them hard to interpret. Hence, we use normalized CPTK (NCPTK) which takes values between 0 and 1, and is calculated as shown in Figure \ref{fig:cptk_example}.

However, CPTKs can not handle labeled edges and were originally applied to dependency trees containing only lexicals. 
%Another solution, proposed by Croce et al. \shortcite{croce-etal-2011-structured} and used in this article, is to include edge labels, i.e., DEPREL, as separate nodes. 
In this article, we use an extension proposed by Croce et al. \shortcite{croce-etal-2011-structured}, which includes edge labels (DEPREL) as separate nodes. The resulting computational structure, the Grammatical Relation Centered Tree (GRCT), is illustrated in Figure~\ref{fig:grct_example}. A dependency tree is transformed into a GRCT by making each UPOS node a child of a DEPREL node and a father of a FORM node. 

\section{Method}
\label{sec:method}
To explore the consistency of errors while parsing numerals, we have used UD treebanks for 4 European languages (2 Germanic and 2 Slavic). To simplify, we considered only sentences containing numerals representing years, later referred to as \emph{original sentences}. We defined these numerals as 4 digits surrounded by spaces, via the simple regular expression {\tt "(?<= )\textbackslash d\{4\}(?= )"}. We then sampled uniformly at random 50 integers between 1100 and 2100 using a fixed random seed, and replaced the occurrences of the previously identified numerals in the original sentences by each of these numbers. Thus, for every found original sentence in a treebank, we synthesized 50 \emph{augmented sentences} (later referred to as \emph{an augmented batch}), only differing in the 4-digit numbers. 
%We only substituted one number per sentence, namely the first found occurrence of any number from the sampled 50 integers. 
We only substituted the first found occurrence of a 4-digit number in a sentence. However, if the same number appeared multiple times in the sentence, then all its occurrences were substituted.

Given such minor changes, a consistent dependency parser should output the same dependency tree for every sentence in each augmented batch. These trees should not necessarily be the same as gold original trees (although this is obviously desirable), but at the very least, the errors made in each augmented batch should be of the same kind. We consider two trees to have the errors of the same kind, and thus belonging to the same \emph{cluster of errors}, if their dependency trees only differ in the 4-digit numerals. All DEPRELs, UPOS tags and FEATS should be exactly the same for any two trees in the same cluster.

Evidently, not all 4-digit numbers in the original sentences were actually years, but the argument about the consistency of errors still stands even if the numbers were amounts of money, temperatures, etc. The magnitude of the numbers was not drastically changed (they are still 4-digit numbers), so the sentences should remain intelligible also after substitution.

In order to evaluate both the consistency of errors and correctness of a dependency parser after introducing the changes above, we need to answer the following questions.
\begin{enumerate}[label=Q\arabic*]
    \item How many augmented batches are parsed completely correctly?
    \begin{itemize}
        \item if the corresponding original sentence is parsed correctly
        \item if the corresponding original sentence is parsed incorrectly
    \end{itemize}
    \item How many sentences in each augmented batch are parsed correctly on average?
    \begin{itemize}
        \item if the corresponding original sentence is parsed correctly
        \item if the corresponding original sentence is parsed incorrectly
    \end{itemize}
    \item How many augmented batches corresponding to incorrectly parsed original sentences have consistent errors, i.e. have the same dependency trees within a batch except FORMs and LEMMAs?
    \item On average, how many clusters of errors does an augmented batch with inconsistent errors have?
    \item On average, how similar are dependency trees in the clusters found in Q4?
\end{enumerate}

Answering Q1 to Q3 is trivial by parsing original and augmented sentences using a pre-trained dependency parser and calculating descriptive statistics. To answer Q4 and Q5, we propose to calculate NCPTK for each pair of trees in an augmented batch. To perform the calculations, we transform each dependency tree to GRCT replacing FORMs (which will be different by experimental design) with the FEATS. We can then construct an undirected graph, where each node is a dependency tree in the batch and two nodes are connected if their NCPTK is exactly 1 (i.e., their dependency trees are identical). Then the problem of finding error clusters in Q4 boils down to finding all maximal cliques in the induced undirected graph, for which we use Bron–Kerbosch algorithm \cite{bron1973algorithm}. Similarity of dependency trees in the given clusters can be assessed using the already calculated NCPTKs, which will provide the answer to Q5.

In hopes of improving parsers' performance and consistency of errors we have also tried to retrain the tokenizer, lemmatizer, PoS tagger and dependency parser (later referred to as a \emph{pipeline}) from scratch using two approaches. The first approach relies on \emph{numeral augmentation} and starts by sampling 20 four-digit integers using a different random seed (while ensuring no overlap with the previously used 50 integers). Using these 20 new numbers and the same procedure as before, we synthesized 20 additional sentences per each previously found original sentence in the training and development treebanks. We will refer to treebanks formed by original and newly synthesized sentences as \emph{augmented treebanks}. The second approach uses \emph{token substitution} and replaces previously found four-digit integers with a special token {\tt NNNN}. The training and development treebanks after this procedure keep their size the same (in constrast to the numeral augmentation method) and will be later referred to as \emph{substituted treebanks}.

We have used Stanza \cite{qi-etal-2020-stanza} to get pretrained dependency parsers as well as to train the whole pipeline from scratch and UDon2 \cite{kalpakchi-boye-2020-udon2} to perform the necessary manipulations on dependency trees and calculate NCPTK. The code is available at \href{https://github.com/dkalpakchi/ud_parser_consistency}{{\tt https://github.com/dkalpakchi/ud\_parser\_consistency}}.

\section{Experimental results}
\subsection{Pretrained pipeline}
We have started the experiment by parsing all original and augmented sentences in the training and development treebanks of the respective languages. The results summary for the off-the-shelf parser are presented in Table~\ref{tab:pretrained_desc}. To our surprise, some sentences were not segmented correctly, i.e. one sentence became multiple, both among original and augmented sentences. However, we did not find any consistent pattern: for instance, the Swedish parser made more segmentation errors for augmented sentences, whereas all the other parsers exhibited the opposite. Nonetheless, we have excluded the cases with wrong sentence segmentation from further analysis. The final number of sentences considered is shown in the rows ``Original considered'' and ``Augmented considered'' in Table~\ref{tab:pretrained_desc}.

\begin{table}[h]
\centering
\begin{tabular}{|l|c|c|c|c|c|c|c|c|}
\hline
\multirow{2}{*}{\bf Metric} & \multicolumn{2}{c|}{\bf English} & \multicolumn{2}{c|}{\bf Swedish} & \multicolumn{2}{c|}{\bf Russian} & \multicolumn{2}{c|}{\bf Ukrainian} \\ \cline{2-9} 
& Train & Dev & Train & Dev & Train & Dev & Train & Dev \\ \hline
Original in total & 235 & 14 & 108 & 5 & 1420 & 270 & 103 & 29\\ 
Wrong sent. segm. & 12 & 0 & 2 & 0 & 25 & 5 & 1 & 1 \\
Original considered & 223 & 14 & 106 & 5 & 1395 & 265 & 102 & 28 \\
Corr. parsed sent. & 53 & 1 & 76 & 1 & 360 & 53 & 27 & 2 \\
Corr. parsed sent. (\%) & 23.8\% & 7.1\% & 71.7\% & 20\% & 25.8\% & 20\% & 26.5\% & 7.1\% \\ \hline
Augmented in total & 11150 & 700 & 5300 & 250 & 69750 & 13250 & 5100 & 1400 \\ 
Wrong sent. segm. & 0 & 0 & 17 & 14 & 13 & 0 & 0 & 0 \\
Augmented considered & 11150 & 700 & 5283 & 236 & 69737 & 13250 & 5100 & 1400 \\
Corr. parsed sent. & 2689 & 50 & 3525 & 43 & 17787 & 2540 & 1227 & 100 \\
Corr. parsed sent. (\%) & 24.1\% & 7.1\% & 66.7\% & 18.2\% & 25.5\% & 19.2\% & 24.1\% & 7.1\% \\ \hline
\end{tabular}
\caption{Results of parsing the original and augmented sentences with pre-trained parsers from Stanza. ``Corr'' stands for ``Correctly'', ``sent'' stands for sentence(s)}
\label{tab:pretrained_desc}
\end{table}

%We have excluded metrics commonly used within UD community, e.g. UAS, LAS or BLEX, because we did not expect such a minor sentence modification to change the attachment of errors drastically. Indeed, we have observed only minor changes (less than 1 unit) for these metrics are thus omit them in this report. 
We have excluded metrics commonly used within UD community, e.g.\ UAS, LAS or BLEX, because for these metrics we observed only minor changes (less than 1 percentage point).  
Another argument for omitting these metrics is that while they are useful in comparing different parsers, they do not fully reflect the usefulness of the parsers in downstream applications. In fact, even a minor error in attaching one dependency arc might lead to a completely wrong tree for the task at hand (depending on how close the error is to the root). Keeping this in mind, we compared accuracy on the sentence level only (reported in the rows ``Correctly parsed'' in Table \ref{tab:pretrained_desc}). We deemed a sentence to be correctly parsed if the NCPTK between its dependency tree and its gold counterpart was 1. We transformed all trees to GRCT and replaced FORM with FEATS, thus requiring not only all DEPREL to be identical, but also all UPOS and FEATS. As can be seen, the number of correctly parsed sentences is either on par or worse for augmented sentences, reaching a performance drop of 5 percentage points for the Swedish training set!

Results of a more detailed analysis needed for answering questions 1 - 5 (posed in Section \ref{sec:method}) are reported in Tables \ref{tab:pretrained_en} - \ref{tab:pretrained_uk}. We adopt the following notation for these tables: ``Original +'' (``Original -'') indicates cases when the original sentence was correctly (incorrectly) parsed. ``QX'' indicates a row with data necessary for answering question X, ``Corr'' stands for ``Correct(ly)'', ``sent'' stands for sentences.

% The following observations are relevant for this article.
% \begin{enumerate}
%     \item If the original sentences are incorrectly parsed, the vast majority of sentences in the corresponding augmented batches will also be incorrectly parsed.
%     \item The fact that an original sentence is correctly parsed does not mean that all sentences in augmented batches will be correctly parsed. In fact, the number of wrong batches can be surprisingly large, i.e., 24 (31.5\%) for Swedish, 19 (5.3\%) for Russian. 4 (7.5\%) for English and 3 (11.1\%) for Ukrainian.
%     \item The errors in augmented batches are not consistent. The degree of inconsistency differs between the languages ranging from around 17\% for Russian training set to 75\% for Swedish dev set. The average observed inconsistency of errors is around 44\%.
%     \item The degree of inconsistency has a similar magnitude between the training and development sets.
%     \item The most typical number of error clusters is 2 and maximum observed is 10.
%     \item The trees between the error clusters have mostly low NCPTK indicating either a large number of errors or errors occurring early on (close to the root).
% \end{enumerate}

We observe a number of interesting patterns from these reports. If the original sentences are incorrectly parsed, the vast majority of sentences in the corresponding augmented batches will also be incorrectly parsed (see mean and median in Q2 rows for ``Original -''). The fact that an original sentence is correctly parsed does not mean that all sentences in augmented batches will be correctly parsed (see mean and median in Q2 rows for ``Original +''). In fact, the number of wrong batches in such a case can be surprisingly large, e.g.\ 24 (31.5\%) for the Swedish training set. 

\begin{table}[H]
\centering
\begin{tabular}{|l|c|c|c|c|}
\hline \multirow{2}{*}{\bf Metric} & \multicolumn{2}{c|}{\bf Training set} & \multicolumn{2}{c|}{\bf Development set} \\ \cline{2-5}
& Original + & Original - & Original + & Original -\\ \hline
Batches considered & 53 & 170 & 1 & 13\\
Completely corr. batches (Q1) & 49 & 0 & 1 & 0 \\
\hline
Corr. parsed sent. within a batch (Q2) & & & & \\
\FirstIndent Mean (SD) & 49 (6.14) & 0.54 (3.67) & 50 (0) & 0 (0)\\
\FirstIndent Median (Min - Max) & 50 (5 - 50) & 0 (0 - 37) & 50 (50 - 50) & 0 (0 - 0)\\
\hline
Batches with consistent errors (Q3) & 0 & 101 & NA & 4\\
\hline
Number of error clusters (Q4) &  & & &\\
\FirstIndent Mean (SD) & 2 (0) & 2.63 (0.95) & NA & 3.89 (2.64)\\
\FirstIndent Median (Min - Max) & 2 (2 - 2) & 2 (2 - 7) & NA & 3 (2 - 10)\\
\hline
Between-cluster NCPTK (Q5) & & & &\\
\FirstIndent Mean (SD) & 0 (0) & 0.07 (0.15) & NA & 0.04 (0.09)\\
\FirstIndent Median (Min - Max) & 0 (0 - 0) & 0 (0 - 0.8) & NA & 0 (0 - 0.28)\\
\hline
\end{tabular}
\caption{A detailed analysis of the parsing results for English using a pretrained pipeline}
\label{tab:pretrained_en}
\end{table}

\begin{table}[H]
\centering
\begin{tabular}{|l|c|c|c|c|}
\hline \multirow{2}{*}{\bf Metric} & \multicolumn{2}{c|}{\bf Training set} & \multicolumn{2}{c|}{\bf Development set} \\ \cline{2-5}
& Original + & Original - & Original + & Original -\\ \hline
Batches considered & 76 & 30 & 1 & 4\\
Completely corr. batches (Q1) & 52 & 0 & 0 & 0 \\
\hline
Corr. parsed sent. within a batch (Q2) & & & & \\
\FirstIndent Mean (SD) & 45.05 (10.77) & 3.37 (10.5) & 43 (0) & 0 (0)\\
\FirstIndent Median (Min - Max) & 50 (0 - 50) & 0 (0 - 42) & 43 (43 - 43) & 0 (0 - 0)\\
\hline
Batches with consistent errors (Q3) & 0 & 16 & 0 & 1\\
\hline
Number of error clusters (Q4) &  & & &\\
\FirstIndent Mean (SD) & 2.29 (0.68) & 2.43 (1.05) & 2 (0) & 2.33 (0.47)\\
\FirstIndent Median (Min - Max) & 2 (2 - 4) & 2 (2 - 5) & 2 (2 - 2) & 2 (2 - 3)\\
\hline
Between-cluster NCPTK (Q5) & & & &\\
\FirstIndent Mean (SD) & 0.04 (0.12) & 0.04 (0.11) & 0 (0) & 0.0002 (0.0003)\\
\FirstIndent Median (Min - Max) & 0 (0 - 0.67) & 0 (0 - 0.37) & 0 (0 - 0) & 0 (0 - 0.0008)\\
\hline
\end{tabular}
\caption{A detailed analysis of the parsing results for Swedish using a pretrained pipeline}
\label{tab:pretrained_sv}
\end{table}

\begin{table}[H]
\centering
\begin{tabular}{|l|c|c|c|c|}
\hline \multirow{2}{*}{\bf Metric} & \multicolumn{2}{c|}{\bf Training set} & \multicolumn{2}{c|}{\bf Development set} \\ \cline{2-5}
& Original + & Original - & Original + & Original -\\ \hline
Batches considered & 360 & 1035 & 53 & 212\\
Completely corr. batches (Q1) & 341 & 0 & 48 & 0 \\
\hline
Corr. parsed sent. within a batch (Q2) & & & & \\
\FirstIndent Mean (SD) & 48.85 (6.34) & 0.19 (2.11) & 47.87 (7.81) & 0.01 (0.21)\\
\FirstIndent Median (Min - Max) & 50 (2 - 50) & 0 (0 - 41) & 50 (3 - 50) & 0 (0 - 3)\\
\hline
Batches with consistent errors (Q3) & 0 & 860 & 0 & 173\\
\hline
Number of error clusters (Q4) &  & & &\\
\FirstIndent Mean (SD) & 2.21 (0.69) & 2.16 (0.43) & 2.2 (0.4) & 2.13 (0.4)\\
\FirstIndent Median (Min - Max) & 2 (2 - 5) & 2 (2 - 4) & 2 (2 - 3) & 2 (2 - 4)\\
\hline
Between-cluster NCPTK (Q5) & & & &\\
\FirstIndent Mean (SD) & 0.08 (0.18) & 0.04 (0.14) & 0 (0) & 0.08 (0.2)\\
\FirstIndent Median (Min - Max) & 0 (0 - 0.67) & 0 (0 - 0.75) & 0 (0 - 0) & 0 (0 - 0.72)\\
\hline
\end{tabular}
\caption{A detailed analysis of the parsing results for Russian using a pretrained pipeline}
\label{tab:pretrained_ru}
\end{table}

\begin{table}[H]
\centering
\begin{tabular}{|l|c|c|c|c|}
\hline \multirow{2}{*}{\bf Metric} & \multicolumn{2}{c|}{\bf Training set} & \multicolumn{2}{c|}{\bf Development set} \\ \cline{2-5}
& Original + & Original - & Original + & Original -\\ \hline
Batches considered & 27 & 75 & 2 & 26\\
Completely corr. batches (Q1) & 24 & 0 & 2 & 0 \\
\hline
Corr. parsed sent. within a batch (Q2) & & & & \\
\FirstIndent Mean (SD) & 45.41 (13.14) & 0.01 (0.11) & 50 (0) & 0 (0)\\
\FirstIndent Median (Min - Max) & 50 (4 - 50) & 0 (0 - 1) & 50 (50 - 50) & 0 (0 - 0)\\
\hline
Batches with consistent errors (Q3) & 0 & 52 & NA & 11\\
\hline
Number of error clusters (Q4) &  & & &\\
\FirstIndent Mean (SD) & 2 (0) & 2.61 (1.37) & NA & 2.8 (0.9)\\
\FirstIndent Median (Min - Max) & 2 (2 - 2) & 2 (2 - 8) & NA & 3 (2 - 5)\\
\hline
Between-cluster NCPTK (Q5) & & & &\\
\FirstIndent Mean (SD) & 0 (0) & 0.12 (0.22) & NA & 0.06 (0.19)\\
\FirstIndent Median (Min - Max) & 0 (0 - 0) & 0 (0 - 0.775) & NA & 0 (0 - 0.77)\\
\hline
\end{tabular}
\caption{A detailed analysis of the parsing results for Ukrainian using a pretrained pipeline}
\label{tab:pretrained_uk}
\end{table}

The errors in augmented batches are not consistent. The degree of inconsistency varies between the languages ranging from around 17\% (175 of 1035) for the Russian training set to 75\% (3 of 4) for the Swedish development set (see Q3 rows). The average observed inconsistency of errors is around 44\%. The degree of inconsistency has a similar magnitude between the training and development sets. The most typical number of error clusters is 2 and maximum observed is 10 (see Q4 rows). The trees between the error clusters have mostly low NCPTK (see Q5 rows) indicating either a large number of errors or errors occurring early on (close to the root). We provide some examples of batches with inconsistent errors in the Appendix.

\subsection{Pipeline trained from scratch on treebanks with numeral augmentation}
We have repeated the same experiment as in the previous section, but with a pipeline trained from scratch on augmented treebanks (as outlined in Section \ref{sec:method}). The results summary is reported in Table~\ref{tab:retrained_desc}.

\begin{table}[h]
\centering
\begin{tabular}{|l|c|c|c|c|c|c|c|c|}
\hline
\multirow{2}{*}{\bf Metric} & \multicolumn{2}{c|}{\bf English} & \multicolumn{2}{c|}{\bf Swedish} & \multicolumn{2}{c|}{\bf Russian} & \multicolumn{2}{c|}{\bf Ukrainian} \\ \cline{2-9} 
& Train & Dev & Train & Dev & Train & Dev & Train & Dev \\ \hline
Original in total & 235 & 14 & 108 & 5 & 1420 & 270 & 103 & 29\\ 
Wrong sent. segm. & 5 & 0 & 3 & 0 & 18 & 5 & 0 & 0 \\
Original considered & 230 & 14 & 105 & 5 & 1402 & 265 & 103 & 29 \\
Corr. parsed sent. & 230 & 0 & 97 & 2 & 976 & 48 & 102 & 3 \\
Corr. parsed sent. (\%) & \textbf{100\%} & 0\% & \textbf{92.4\%} & \textbf{40\%} & \textbf{69.6\%} & 18.1\% & \textbf{99\%} & \textbf{10.3\%} \\ \hline
Augmented in total & 11500 & 700 & 5250 & 250 & 70100 & 13250 & 5150 & 1450 \\ 
Wrong sent. segm. & 0 & 0 & 0 & 0 & 13 & 0 & 0 & 0 \\
Augmented considered & 11500 & 700 & 5250 & 250 & 70087 & 13250 & 5150 & 1450 \\
Corr. parsed sent. & 11452 & 0 & 4864 & 100 & 49005 & 2437 & 5100 & 133 \\
Corr. parsed sent. (\%) & \textbf{99.6\%} & 0\% & \textbf{92.7\%} & \textbf{40\%} & \textbf{69.9\%} & 18.4\% & \textbf{99\%} & \textbf{9.2\%} \\ \hline
\end{tabular}
\caption{Results of parsing the original and augmented sentences with the pipeline trained on augmented treebanks. ``Corr'' stands for ``Correctly'', ``sent'' stands for sentence(s). Performance improvements with respect to the pre-trained parser (see Table \ref{tab:pretrained_desc}) are indicated in \textbf{bold}.}
\label{tab:retrained_desc}
\end{table}

Retraining with numeral augmentation resulted in a clear and substantial performance boost for all languages, especially for the training treebanks. Performance boost on the development treebanks is less pronounced and sometimes leads to a slight performance degradation. We attribute this to a possible overfitting, indicating that 20 samples per an original sentence might have been too many and the procedure needs to be refined in future. Nevertheless, the detailed analysis, reported in Appendix, shows that the number of wrong sentence segmentations decreased for all languages and a consistency of errors is either better or on par with the pretrained counterparts. The number of error clusters got reduced to a maximum of 4 compared to 10 for the off-the-shelf parser.

\subsection{Pipeline trained from scratch on treebanks with token substitution}
We have repeated the same experiment as in the previous section, but with a pipeline trained from scratch on substituted treebanks (as outlined in Section \ref{sec:method}). The results summary is reported in Table~\ref{tab:retrained_tokens_desc}.

\begin{table}[h]
\centering
\begin{tabular}{|l|c|c|c|c|c|c|c|c|}
\hline
\multirow{2}{*}{\bf Metric} & \multicolumn{2}{c|}{\bf English} & \multicolumn{2}{c|}{\bf Swedish} & \multicolumn{2}{c|}{\bf Russian} & \multicolumn{2}{c|}{\bf Ukrainian} \\ \cline{2-9} 
& Train & Dev & Train & Dev & Train & Dev & Train & Dev \\ \hline
Substituted in total & 235 & 14 & 108 & 5 & 1420 & 270 & 103 & 29\\ 
Wrong sent. segm. & 14 & 0 & 1 & 0 & 10 & 1 & 2 & 1 \\
Substituted considered & 221 & 14 & 107 & 5 & 1410 & 269 & 101 & 28 \\
Corr. parsed sent. & 81 & 1 & 73 & 2 & 341 & 59 & 23 & 2 \\
Corr. parsed sent. (\%) & \textbf{36.7\%} & 7.1\% & 68.2\% & \textbf{40\%} & 24.2\% & \textbf{21.9\%} & 22.8\% & 7.1\%  \\ \hline
\end{tabular}
\caption{Results of parsing the substituted sentences with the pipeline trained on treebanks with token susbtitution. ``Corr'' stands for ``Correctly'', ``sent'' stands for sentence(s). Performance improvements with respect to the pre-trained parser (see Table \ref{tab:pretrained_desc}) are indicated in \textbf{bold}.}
\label{tab:retrained_tokens_desc}
\end{table}

Retraining with token substitution resulted in a slight performance boost for Russian and Swedish on the development treebanks and a slight performance degradation on the training treebanks for all languages except English. Interestingly, more sentences have been segmented correctly for Russian and Swedish, while the parsers for English and Ukrainian produce more segmentation errors compared to pre-trained parsers. At the same time, more sentences have been segmented incorrectly compared to the numeral augmentation method (except for Russian). Given that all models were re-trained with the same default seed from Stanza, we are unsure what this can be attributed to, other than the choice of the token {\tt NNNN} itself. The tokenization model in Stanza is based on unit (character) embeddings, so a tokenization model might benefit from a token without letters or just from replacing all 4-digit numerals with one fixed integer, say 0000. This is, however, highly speculative and requires further investigation. 

An obvious advantage of token substitution is that the errors become consistent (since no clusters of errors could potentially be formed). However, the observed effect on performance suggests that token substitution with this specific token {\tt NNNN} is not the best solution to the problem.

\section{Conclusion}
We have observed that such a minor change as changing one 4-digit number for another leads to surprising performance fluctuations for pretrained parsers. Furthermore, we have noted the errors to be inconsistent, making the development of downstream applications more complicated. To alleviate the issue we tried out two methods and trained two proof-of-concept pipelines from scratch. One of the methods, namely the numeral augmentation scheme, resulted in substantial performance gains.

Finally, the results of the experiment suggest that UD treebanks might be biased towards specific time intervals, e.g.\ the 19th and 20th centuries. Bias in the data leads to bias in the models making it harder to use the parser for some downstream applications, e.g.\ in the history domain. The results of this experiment also prompt a further and more extensive investigation of possible other biases, such as names of geographical entities, gender pronouns, currencies, etc. 

\section*{Acknowledgements}
This work was supported by Vinnova (Sweden's Innovation Agency) within the project 2019-02997. We would like to thank the anonymous reviewers for their comments and the suggestion to try token substitution.

\bibliographystyle{acl}
\bibliography{anthology,coling2020}

\begin{appendices}
\section{Details of the experimental setup}
We have experimented with the training and development sets of the following treebanks: UD\_English-EWT, UD\_Swedish-Talbanken, UD\_Russian-SynTagRus, UD\_Ukrainian-IU. For sampling 50 integers used for validating the parser's performance, we have seeded Numpy's random number generator with the 1000th prime number (7919). For sampling 20 integers used for augmenting treebanks for re-training, we chosen the 999th prime number (7907) as the random seed. Then we sampled 100 integers, filtered out all overlapping with the previously sampled 50 and then taken the first 20 integers of the remainder.

\section{Detailed results for the pipeline trained from scratch}
\begin{table}[H]
\centering
\begin{tabular}{|l|c|c|c|c|}
\hline \multirow{2}{*}{\bf Metric} & \multicolumn{2}{c|}{\bf Training set} & \multicolumn{2}{c|}{\bf Development set} \\ \cline{2-5}
& Original + & Original - & Original + & Original -\\ \hline
Batches considered & 230 & 0 & 0 & 14\\
Completely corr. batches (Q1) & 229 & NA & NA & 0 \\
\hline
Corr. parsed sent. within a batch (Q2) & & & & \\
\FirstIndent Mean (SD) & 49.79 (3.16) & NA & NA & 0 (0)\\
\FirstIndent Median (Min - Max) & 50 (2 - 50) & NA & NA & 0 (0 - 0)\\
\hline
Batches with consistent errors (Q3) & 0 & NA & NA & 4\\
\hline
Number of error clusters (Q4) &  & & &\\
\FirstIndent Mean (SD) & 2 (0) & NA & NA & 2.6 (0.8)\\
\FirstIndent Median (Min - Max) & 2 (2 - 2) & NA & NA & 2 (2 - 4)\\
\hline
Between-cluster NCPTK (Q5) & & & &\\
\FirstIndent Mean (SD) & 0 (0) & NA & NA & 0.05 (0.1)\\
\FirstIndent Median (Min - Max) & 0 (0 - 0) & NA & NA & 0 (0 - 0.31)\\
\hline
\end{tabular}
\caption{A detailed analysis of the parsing results for English using a retrained pipeline}
\label{tab:retrained_en}
\end{table}

\vspace{-1em}

\begin{table}[H]
\centering
\begin{tabular}{|l|c|c|c|c|}
\hline \multirow{2}{*}{\bf Metric} & \multicolumn{2}{c|}{\bf Training set} & \multicolumn{2}{c|}{\bf Development set} \\ \cline{2-5}
& Original + & Original - & Original + & Original -\\ \hline
Batches considered & 97 & 8 & 2 & 3\\
Completely corr. batches (Q1) & 97 & 0 & 2 & 0 \\
\hline
Corr. parsed sent. within a batch (Q2) & & & & \\
\FirstIndent Mean (SD) & 50 (0) & 1.75 (4.63) & 50 (0) & 0 (0)\\
\FirstIndent Median (Min - Max) & 50 (50 - 50) & 0 (0 - 14) & 50 (50 - 50) & 0 (0 - 0)\\
\hline
Batches with consistent errors (Q3) & NA & 7 & NA & 1\\
\hline
Number of error clusters (Q4) &  & & &\\
\FirstIndent Mean (SD) & NA & 3 (0) & NA & 2 (0)\\
\FirstIndent Median (Min - Max) & NA & 3 (3 - 3) & NA & 2 (2 - 2)\\
\hline
Between-cluster NCPTK (Q5) & & & &\\
\FirstIndent Mean (SD) & NA & 0 (0) & NA & 0.04 (0.04)\\
\FirstIndent Median (Min - Max) & NA & 0 (0 - 0) & NA & 0.04 (0 - 0.08)\\
\hline
\end{tabular}
\caption{A detailed analysis of the parsing results for Swedish using a retrained pipeline}
\label{tab:retrained_sv}
\end{table}

\vspace{-1em}

\begin{table}[H]
\centering
\begin{tabular}{|l|c|c|c|c|}
\hline \multirow{2}{*}{\bf Metric} & \multicolumn{2}{c|}{\bf Training set} & \multicolumn{2}{c|}{\bf Development set} \\ \cline{2-5}
& Original + & Original - & Original + & Original -\\ \hline
Batches considered & 976 & 426 & 48 & 217\\
Completely corr. batches (Q1) & 950 & 1 & 44 & 0 \\
\hline
Corr. parsed sent. within a batch (Q2) & & & & \\
\FirstIndent Mean (SD) & 49.58 (3.63) & 1.44 (7.75) & 49.77 (0.92) & 0.22 (2.92)\\
\FirstIndent Median (Min - Max) & 50 (2 - 50) & 0 (0 - 50) & 50 (45 - 50) & 0 (0 - 43)\\
\hline
Batches with consistent errors (Q3) & 0 & 369 & 0 & 149\\
\hline
Number of error clusters (Q4) &  & & &\\
\FirstIndent Mean (SD) & 2.08 (0.27) & 2.09 (0.34) & 2 (0) & 2.13 (0.4)\\
\FirstIndent Median (Min - Max) & 2 (2 - 3) & 2 (2 - 4) & 2 (2 - 2) & 2 (2 - 4)\\
\hline
Between-cluster NCPTK (Q5) & & & &\\
\FirstIndent Mean (SD) & 0.05 (0.14) & 0.08 (0.18) & 0.13 (0.22) & 0.07 (0.2)\\
\FirstIndent Median (Min - Max) & 0 (0 - 0.5) & 0 (0 - 0.67) & 0.003 (0 - 0.5) & 0 (0 - 0.87)\\
\hline
\end{tabular}
\caption{A detailed analysis of the parsing results for Russian using a retrained pipeline}
\label{tab:retrained_ru}
\end{table}

\begin{table}[H]
\centering
\begin{tabular}{|l|c|c|c|c|}
\hline \multirow{2}{*}{\bf Metric} & \multicolumn{2}{c|}{\bf Training set} & \multicolumn{2}{c|}{\bf Development set} \\ \cline{2-5}
& Original + & Original - & Original + & Original -\\ \hline
Batches considered & 102 & 1 & 3 & 26\\
Completely corr. batches (Q1) & 102 & 0 & 2 & 0 \\
\hline
Corr. parsed sent. within a batch (Q2) & & & & \\
\FirstIndent Mean (SD) & 50 (0) & 0 (0) & 44.33 (8.01) & 0 (0)\\
\FirstIndent Median (Min - Max) & 50 (50 - 50) & 0 (0 - 0) & 50 (33 - 50) & 0 (0 - 0)\\
\hline
Batches with consistent errors (Q3) & NA & 1 & 0 & 13\\
\hline
Number of error clusters (Q4) &  & & &\\
\FirstIndent Mean (SD) & NA & NA & 2 (0) & 2.46 (0.75)\\
\FirstIndent Median (Min - Max) & NA & NA & 2 (2 - 2) & 2 (2 - 4)\\
\hline
Between-cluster NCPTK (Q5) & & & &\\
\FirstIndent Mean (SD) & NA & NA & 0.29 (0) & 0.09 (0.22)\\
\FirstIndent Median (Min - Max) & NA & NA & 0.29 (0.29 - 0.29) & 0 (0 - 0.67)\\
\hline
\end{tabular}
\caption{A detailed analysis of the parsing results for Ukrainian using a retrained pipeline}
\label{tab:retrained_uk}
\end{table}

\section{Examples of batches with inconsistent errors}
In this section we report dependency trees from the augmented batch with the largest observed number of error clusters (which happened to be 10 clusters for the English development set). The original sentences in these clusters were too long, so we have pruned the dependency trees to include only the differing subtrees. The cluster sizes and included numerals are as follows:
\begin{enumerate}[label=Cluster \arabic*.,leftmargin=6em]
    \item 2 trees (numerals 1505, 1505)
    \item 3 trees (numerals 1798, 1777, 1817)
    \item 3 trees (numerals 1872, 1844, 1883)
    \item 3 trees (numerals 1361, 1338, 1427)
    \item 4 trees (numerals 1704, 1605, 1662, 1562)
    \item 5 trees (numerals 1420, 1344, 1295, 1504, 1299)
    \item 5 trees (numerals 1625, 1599, 1564, 1564, 1493)
    \item 6 trees (numerals 1128, 2024, 1147, 1182, 2030, 1205)
    \item 7 trees (numerals 1964, 1308, 1415, 1413, 1404, 1967, 1413)
    \item 8 trees (numerals 1774, 1721, 1759, 1759, 1461, 1731, 1724, 1832)
\end{enumerate}

\foreach \n in {1,...,10} {
    \begin{figure}[H]
        \centering
        \includegraphics[width=\textwidth]{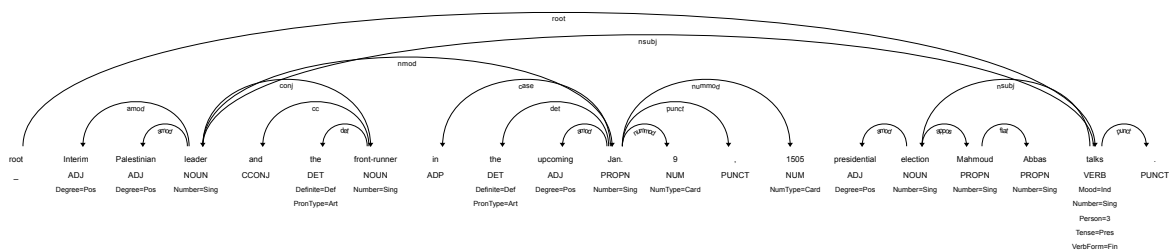}
    	\caption{An example truncated dependency tree from cluster \n}
    \end{figure}
}

\end{appendices}

\end{document}